\pdfoutput=1

\documentclass[11pt]{article}
\usepackage[]{authblk}
\usepackage[]{acl}

\usepackage{times}
\usepackage{latexsym}

\usepackage[T1]{fontenc}

\usepackage[utf8]{inputenc}

\usepackage{microtype}

\usepackage{multirow}
\usepackage{microtype}
\usepackage{times}
\usepackage{latexsym}
\usepackage{microtype}
\usepackage{graphicx}
\usepackage{multirow}
\usepackage{soul}
\usepackage{amsmath}
\usepackage{bbm}
\usepackage{amssymb}
\usepackage{pifont}
\newcommand{\cmark}{\ding{51}}%
\newcommand{\xmark}{\ding{55}}%

\usepackage{enumitem}
\usepackage{setspace}
\usepackage[belowskip=0pt,aboveskip=4pt]{caption}

\usepackage{tablefootnote}
\usepackage{booktabs}
\usepackage{arydshln}
\usepackage{xspace}
\usepackage{color}
\usepackage[inline,ignoremode]{trackchanges}
\addeditor{jianshu}
\addeditor{wenlin}

\usepackage{pifont}

\newcommand{\annotate}[1]{{$#1^{\scalebox{0.7}{\ding{61}}}$}}

\newcommand{\model}{\textbf{\texttt{C-MORE}}\xspace}

\title{C-MORE: Pretraining to Answer Open-Domain Questions by \\ Consulting Millions of References}

\author[1,\thanks{\ \ Work was done when interning at Tencent AI Lab.}]{Xiang Yue}
\author[2]{Xiaoman Pan}
\author[2]{Wenlin Yao}
\author[2]{Dian Yu}
\author[2]{Dong Yu}
\author[2]{Jianshu Chen}

\affil[1]{The Ohio State University}
\affil[2]{Tencent AI Lab}
\affil[ ]{\texttt{yue.149@osu.edu}}
\affil[ ]{\{\texttt{xiaomanpan,wenlinyao,yudian,dyu,jianshuchen\}@tencent.com}}

\begin{document}
\maketitle

\begin{abstract}
We consider the problem of pretraining a two-stage open-domain question answering (QA) system (retriever + reader) with strong transfer capabilities. The key challenge is how to construct a large amount of high-quality question-answer-context triplets without task-specific annotations. Specifically, the triplets should align well with downstream tasks by: (i) covering a wide range of domains (for open-domain applications), (ii) linking a question to its semantically relevant context with supporting evidence (for training the retriever), and (iii) identifying the correct answer in the context (for training the reader). Previous pretraining approaches generally fall short of one or more of these requirements. In this work, we automatically construct a large-scale corpus that meets all three criteria by consulting millions of references cited within Wikipedia. The well-aligned pretraining signals benefit both the retriever and the reader significantly. Our pretrained retriever leads to 2\%-10\% absolute gains in top-20 accuracy. And with our pretrained reader, the entire system improves by up to 4\% in exact match.\footnote{Our code, data, and pretrained models are available at: \url{https://github.com/xiangyue9607/C-MORE}}

\end{abstract}
\section{Introduction}
Open-domain question answering (QA) aims to extract the answer to a question from a large set of passages. A simple yet powerful approach adopts a two-stage framework \cite{chen2017reading,karpukhin2020dense}, which first employs a retriever to fetch a small subset of relevant passages from large corpora (i.e., \textit{retriever}) and then feeds them into a \textit{reader} to extract an answer (text span) from them. Due to its simplicity, a sparse retriever such as TF-IDF/BM25 is generally used together with a trainable reader \cite{min2019discrete}. However, recent advances show that transformer-based dense retrievers trained on supervised data \cite{karpukhin2020dense} can greatly boost the performance, which better captures the semantic relevance between the question and the correct passages. Such approaches, albeit promising, are restricted by the limited amount of human annotated training data.

Inspired by the recent progresses of language models pretraining \cite{Devlin19Bert,lee2019latent,guu2020realm,SachanPSKPHC20}, we would like to address the following central question: \textit{can we pretrain a two-stage open-domain QA system (retriever + reader) without task-specific human annotations?} Unlike general language models, pretraining such a system that has strong transfer capabilities to downstream open-domain QA tasks is challenging. This is mainly due to the lack of well-aligned pretraining supervision signals. In particular, we need the constructed pretraining dataset (in the form of question-answer-context triplets) to: (i) cover a wide range of domains (for open-domain applications), (ii) link a question to its semantically relevant context with supporting evidence (for training the retriever), and (iii) identify the correct answer in the context (for training the reader).

\begin{figure*}
    \centering
    \includegraphics[width=\linewidth]{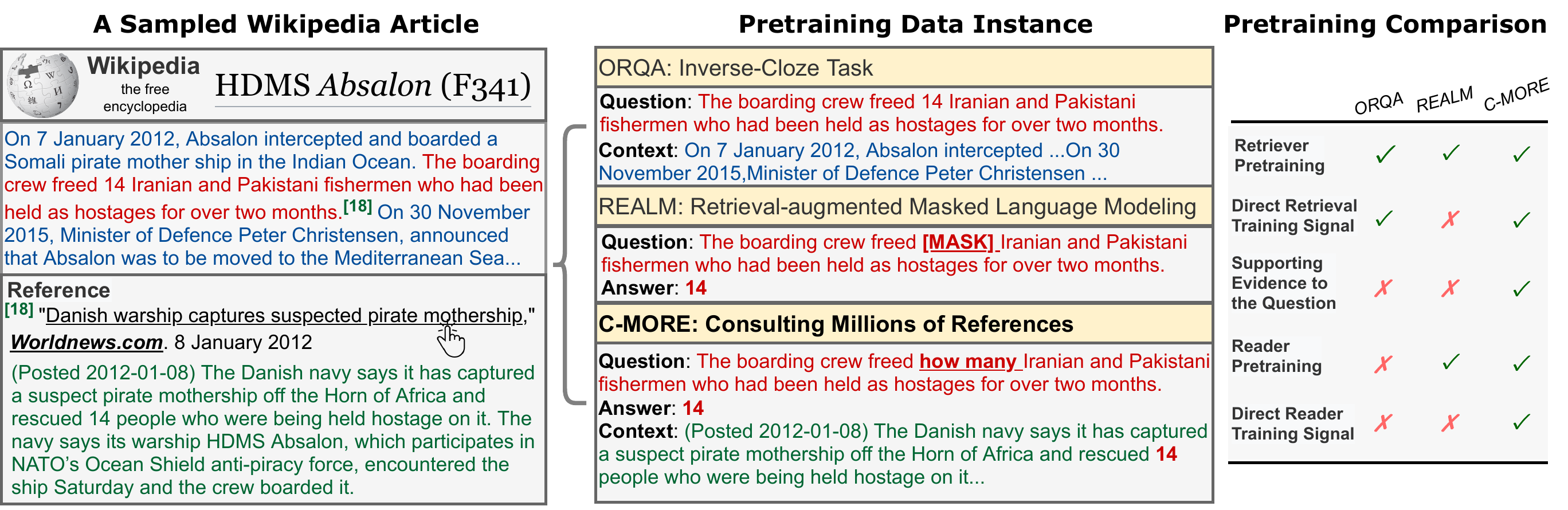}
    \caption{Different pretraining methods for open-domain QA. Our \model pretrains both retriever and reader by using direct signals extracted from millions of references cited in the verified knowledge source.}
    \label{fig:data_collection}
\end{figure*}

There have been several recent attempts in addressing these challenges. ORQA \cite{lee2019latent} creates pseudo query-passage pairs by randomly sampling a sentence from a paragraph and treating the sampled sentence as the question while the rest sentences as the context. REALM \cite{guu2020realm} adopts a retrieve-then-predict approach, where the context is dynamically retrieved during training and an encoder (reader) predicts the masked token in the question based on the retrieved context. The retriever pretraining signals constructed in these approaches are not aligned with question-context pairs in open-domain QA settings. For example, as shown in Figure \ref{fig:data_collection}, the context (in blue color) of ORQA pretraining data instance does not contain direct supporting evidence to the question. Likewise, the dynamically retrieved context in REALM cannot be guaranteed to contain direct supporting evidence either. In addition, existing pretraining methods \cite{lee2019latent,guu2020realm} mostly focus on the retriever and do not jointly provide direct pretraining signals for the reader (Figure \ref{fig:data_collection}).




To meet all three aforementioned criteria, we propose a pretraining approach named \textbf{C}onsulting \textbf{M}illions \textbf{O}f \textbf{RE}ferences (\model), which automatically constructs pretraining data with well-aligned supervision signals (Figure \ref{fig:data_collection}). Specifically, we first extract three million statement-reference pairs from 
Wikipedia along with its cited references. Then, we transform them into question-answer-context triplets by replacing a potential answer span in the statement (e.g., \emph{``14''} in the Figure \ref{fig:data_collection}) by an interrogative phrase (e.g, \emph{``how many''}). Such kind of pseudo triplets are in the exact same form as human-annotated ones, and the question is linked to the context that contains the most direct-supporting evidence, a highly desirable feature for open-domain QA tasks. We experiment the pretraining with a widely-adopted open-domain QA system, Dense Passage Retriever (DPR) \cite{karpukhin2020dense}. The experimental results show that our pretrained retriever not only outperforms both sparse and dense retrieval baselines in the zero-shot retrieval setting (2\%-10\% absolute gain in top-20 accuracy), but also leads to further improvement in the downstream task finetuning. By integrating with our pretrained reader, the entire open-domain pretraining improves the end-to-end QA performance by 4\% in exact match.

\section{Method}
Recall that we want to automatically construct a large-scale open-domain QA pretraining dataset that satisfies three criteria:
(i) The dataset should cover a wide range of domains for the open-domain QA purpose. (ii) The context passage is semantically relevant to the question and contains direct supporting evidence for answering the question. (iii) The correct answer span in the context passage for answering the question should also be identified for training the reader. This section first discusses how to extract a large amount of statement-reference pairs from the Wikipedia and then explain how to construct pseudo question-answer-context triplets for pretraining open-domain QA systems.

\subsection{Statement-Reference Pairs Collection}
Wikipedia articles usually contain a list of knowledge sources (references) at the end that are verified by human editors to support the statements in the articles \cite{li2020harvesting}. And the reference documents always consist of strong supporting evidence to the statements. For example, as shown in Figure \ref{fig:data_collection}, the document (in green color) contains the direct evidence \textit{``...rescued 14 people who were being held hostage on it...''} to support the query (red text) \textit{``The boarding crew freed 14 Iranian and Pakistani fishermen who had been held as hostages over two months''}. Additionally, such knowledge sources are often organized in a good structure and can be automatically extracted and processed.
Moreover, the statement-reference pairs in Wikipedia cover a wide range of topics and domains. Thus, when converted into question-context pairs, they satisfy the first two criteria and are suitable for training an accurate dense retriever at a large scale.

In our study, we extract around six million statement-reference pairs from Wikipedia. We filter the pairs whose reference documents are not reachable and finally obtain around three million statement-reference pairs (see statistics in Appendix Table \ref{tbl:dataset}). 
The data collection method we proposed is very general and therefore can be easily extended to other domains, e.g., WikiEM (\href{wikem.org}{wikem.org}) for medical domain or other languages, e.g., Baidu Baike (\href{baike.baidu.com}{baike.baidu.com}) for Chinese.

\subsection{QAC Triplets Construction}
We now explain how to further convert the statement-reference pairs into question-answer-context pairs.
Inspired by previous unsupervised extractive QA work~\cite{lewis2019unsupervised}, we extract entities as potential answers to construct pseudo question-answer-context pairs where an answer span is extracted from the context given an question to accommodate the extractive QA setting.
Specifically, we first adopt an off-the-shelf named entity recognition tool spaCy~\cite{spacy2} to identify entities in each query. Next, we filter the entities that do not appear in the evidence based on string matching. If multiple entities are found, we sample one of them as the potential answer to the query. The sampled entity in the query is replaced by an interrogative phrase based on the entity type (e.g., a \texttt{[DATE]} entity will be replaced by phrases such as \emph{``when''}, \emph{``what date''}. In this way, we can construct question-answer-context triplets to train open-domain QA models. See more question reformation rules in Appendix Table \ref{tbl:question_word}).

\section{Experiment}
\begin{table}[!t]
\resizebox{\linewidth}{!}{%
\begin{tabular}{l|l|ccc}
\hline
Data Type & Dataset & Train & Dev & Test \\ \hline
Pretraining & \model & 2.96M & 40K & - \\ \hline
\multirow{3}{*}{\begin{tabular}[c]{@{}l@{}}Finetuning \\ QA Data\end{tabular}} &  NaturalQuestion  & 58,880 & 8,757 & 3,610 \\
 & TriviaQA & 60,413 & 8,837 & 11,313 \\
 & WebQuestion & 2,474 & 361 & 2,032 \\ \hline
\end{tabular}
}
\caption{Statistics of pretraining and finetuning data.}
\label{tbl:dataset}
\end{table}
\begin{table*}[t!]
\centering
\small
\resizebox{\linewidth}{!}{%
\begin{tabular}{l|l|l|c|c|c|ccc}
\hline
\multirow{2}{*}{Settings} & \multirow{2}{*}{Methods} & \multirow{2}{*}{\begin{tabular}[c]{@{}l@{}}Training\\ Data\end{tabular}} & \multicolumn{3}{c|}{Top-20 Accuracy} & \multicolumn{3}{c}{Top-100 Accuracy}  \\ \cline{4-9} 
 &  &  & NQ & TQA & WebQ & \multicolumn{1}{c|}{NQ} & \multicolumn{1}{c|}{TQA} & WebQ \\ \hline\hline
\multirow{5}{*}{Unsupervised} & BM25 & - & 59.1* & 66.9* & 55.0* & \multicolumn{1}{c|}{73.7*} & \multicolumn{1}{c|}{76.7*} & 71.1* \\
 & ORQA \cite{lee2019latent} & Wikipedia &\annotate{50.6} & \annotate{57.5} &-  & \multicolumn{1}{c|}{\annotate{66.8}} & \multicolumn{1}{c|}{\annotate{73.6}} &-  \\
 & REALM \cite{guu2020realm} & Wikipedia &\annotate{59.8}  &\annotate{ 68.2}  &-  & \multicolumn{1}{c|}{\annotate{74.9}} & \multicolumn{1}{c|}{\annotate{79.4}} &-  \\
 & \textbf{\model} & \textbf{Wikipedia} & \textbf{61.9} & \textbf{72.2} & \textbf{62.7} & \multicolumn{1}{c|}{\textbf{75.8}} & \multicolumn{1}{c|}{\textbf{81.3}} & \textbf{78.5} \\ \hline\hline
\multirow{6}{*}{\begin{tabular}[c]{@{}l@{}}Domain\\ Aaptation\end{tabular}} & DPR-NQ & NaturalQuestion & - & 69.7 & 69.0 & \multicolumn{1}{c|}{-} & \multicolumn{1}{c|}{79.2} & 78.8 \\
 & \textbf{+ \model} & + Wikipedia & - & \textbf{72.8} & \textbf{71.2} & \multicolumn{1}{c|}{-} & \multicolumn{1}{c|}{\textbf{81.6}} & \textbf{81.3} \\ \cline{2-9} 
 & DPR-TQA & TriviaQA & 69.2 & - & 71.5 & \multicolumn{1}{c|}{80.3} & \multicolumn{1}{c|}{-} & 81.0 \\
 & \textbf{+ \model} & + Wikipedia & \textbf{71.0} & - & \textbf{74.3} & \multicolumn{1}{c|}{\textbf{81.7}} & \multicolumn{1}{c|}{-} & \textbf{83.2} \\ \cline{2-9} 
 & DPR-WebQ & WebQ & 56.1 &66.1  & - & \multicolumn{1}{c|}{70.7} & \multicolumn{1}{c|}{77.6} & - \\
 & \textbf{+ \model} & + Wikipedia & \textbf{67.3} & \textbf{74.2} & - & \multicolumn{1}{c|}{\textbf{79.2}} & \multicolumn{1}{c|}{\textbf{82.6}} & - \\ \hline\hline
\multirow{2}{*}{Supervised} & DPR-supervised & Supervised Data & 78.4* & 79.4* & 73.2* & \multicolumn{1}{c|}{85.4*} & \multicolumn{1}{c|}{85.0*} & 81.4*  \\
 & \textbf{+ \model} & + Wikipedia & \textbf{80.3} & \textbf{81.3} & \textbf{75.0} & \multicolumn{1}{c|}{\textbf{86.7}} & \multicolumn{1}{c|}{\textbf{85.9}} & \textbf{83.2} \\ \hline
\end{tabular}%
}
\caption{Overall retrieval performance of different models. Results marked with ``*'' are from DPR \cite{karpukhin2020dense}, \annotate{``}'' are from \cite{SachanPSKPHC20} and ``-''  means it does not apply to the current setting. }
\label{tbl:retrieval_results}
\end{table*}
\begin{table*}[t!]
\small
\centering
\begin{tabular}{c|l|c|c|c|c|c|c|c}
\hline
\multirow{2}{*}{Row} &\multirow{2}{*}{Model Architecture} &\multicolumn{2}{c|}{Retriever} & \multicolumn{2}{c|}{Reader} & \multirow{2}{*}{NQ} & \multirow{2}{*}{TQA} & \multirow{2}{*}{WebQ} \\ \cline{3-6}
& & Pretrain & Finetune & Pretrain & Finetune &  &  &  \\ \hline
1 & ORQA \cite{lee2019latent} &\cmark & \cmark & \xmark & \cmark & 33.3 & 45.0 & 36.4 \\ 
2 & REALM \cite{guu2020realm} &\cmark & \cmark & \cmark & \cmark & 40.4 & - & 40.7 \\ \hline\hline
3&\multirow{6}{*}{DPR \cite{karpukhin2020dense}}  &\cmark & \xmark & \cmark & \xmark & 11.3 & 24.8 & 4.5 \\ \cline{1-1} \cline{3-9}
4&  &\xmark & \xmark & \xmark & \cmark & 32.6 & 52.4 & 29.9 \\
5&  &\cmark & \xmark & \xmark & \cmark & \textbf{35.3} & \textbf{55.1} & \textbf{32.1} \\ \cline{1-1} \cline{3-9}
6&   &\xmark & \cmark & \xmark & \cmark & 41.5 & 56.8 & 34.6 \\
7&   &\cmark & \cmark & \xmark & \cmark & \textbf{41.9} & 58.6 & 35.6 \\
8&   &\cmark & \cmark & \cmark & \cmark & 41.6  & \textbf{60.3} & \textbf{38.6} \\ \hline
\end{tabular}%
\caption{End-to-end QA performance based on different retrievers and readers. Note that we only test the effectiveness of \model based on the DPR \cite{karpukhin2020dense} model architecture. ORQA and REALM are  listed here as references. The retriever of Row 4 is BM25, which does not involve either pretraining or finetuning.}
\label{tbl:qa_performance}
\end{table*}

\subsection{Experimental Setup}
\noindent\textbf{Pretraining Model Architecture}. Since conceptually the construed triplets is in the same format as the annotated QA data, they can be used to pretrain any existing neural open-domain QA model. Here, we adopt DPR~\cite{karpukhin2020dense}, which consists of a dual-encoder as the retriever and a BERT reader, considering its effectiveness and popularity. Specifically, the retriever first retrieves top-$k$ (up to 400 in our experiment) passages, and the reader assigns a passage score to each retrieved passage and extracts an answer with a span score. The span with the highest passage selection score is regarded as the final answer. The reader and retriever can be instantiated with different models and we use \texttt{BERT-base-uncased} for both of them following~\cite{karpukhin2020dense}.

\noindent\textbf{Pretraining Data Processing}. For our extracted pseudo question-answer-context triplets, sometimes the context (reference document) is too long to fit into a standard BERT (maximum 512 tokens) in the DPR model. 
Thus, we chunk a long document into $n$-word text blocks with a stride of $m$. Without loss of generality, we use multiple combinations of $n$ and $m$: $n=\{128,256,512\}$, $n=\{64,128,256\}$. Then we calculate relevance scores (using BM25) of the derived blocks with the question and select the most relevant block as the context. Note that the retrieval step is done within the single document (usually less than 20 text blocks). In contrast, the baseline model (Section \ref{sec:retrieval_performance}) - sparse retriever BM25 - looks up the entire knowledge corpus (20M text blocks). In this way, we can automatically collect the most relevant context that supports the query from a long article. 



\noindent\textbf{Finetuning QA Datasets.} 
We consider three popular open-domain QA datasets for finetuning: NaturalQuestions (NQ) \cite{kwiatkowski2019natural}, TriviaQA (TQA) \cite{joshi2017triviaqa}, and WebQuestions (WebQ) \cite{berant2013semantic}, whose statistics are shown in Table \ref{tbl:dataset}. 

Following the setting of DPR \cite{karpukhin2020dense}, we use the Wikipedia as the knowledge source and split Wikipedia articles into 100-word units for retrieval.  All the datasets we use are the processed versions from the DPR implementation.

\noindent\textbf{Overlap between Pretraining and Finetuning Datasets}. Though both \model and downstream QA data are constructed based on Wikipedia, the overlap between them would be very little. \model extracts queries from Wikipedia while the queries of downstream QA data are annotated by human.  \model extracts contexts from the external referenced pages (general Web) while the downstream QA data extract contexts from Wikipedia.

\noindent\textbf{Implementation Details.} 
For pretraining, we set training epochs to 3, batch size to 56 for retrievers and 16 for readers, and learning rate to 2e-5. We select the best checkpoint based on the pretraining dev set. For finetuning, we use the same set of hyperparameters as the original DPR paper. The comparing baselines ORQA \cite{lee2019latent} and REALM \cite{guu2020realm} use 288-token truncation over Wikipedia, which are not directly comparable to our results. To enable a fair comparison, we report the retrieval results from a recent paper \cite{SachanPSKPHC20}, which uses the same retrieval corpus as ours.

\subsection{Retrieval Performance}
\label{sec:retrieval_performance}
We consider three settings to demonstrate the usefulness of our pretrained retriever.

\noindent \textbf{Unsupervised.} We assume no annotated training QA pairs are available. In this setting, We compare our method with existing unsupervised retrievers: a sparse retriever BM25 and two pretrained dense retrievers ORQA and REALM. 

\noindent \textbf{Domain Adaptation.} We consider the condition in which there are QA training pairs in the source domain but no training data in the target domain. The task is to obtain good retrieval performance on the target test set only using source training data. We compare our method with two baselines: one is to directly train a dense retriever on the source domain while the other is to first pretrain a dense retriever on our constructed corpus and then finetune it on the source domain training set.

\noindent \textbf{Supervised.} In this setting, all the annotated QA training instances are used. Similar to the previous setting, we compare a supervised retriever with and without our \model pretraining.

For all settings, we report the top-$k$ retrieval accuracy ($k\in\{20,100\}$) on the test set following \cite{karpukhin2020dense}. See the overall retrieval performance of different models in each setting in Table \ref{tbl:retrieval_results}. We have the following observations.

In the \textbf{unsupervised} setting, compared with the strong sparse retrieval baseline BM25, our pretrained dense retriever shows significant improvement. For example, we obtain around 7\% absolute improvement in terms of both Top-20 and Top-100 accuracy on the WebQuestion dataset. Compared with pretrained dense retrievers (i.e., ORQA and REALM), our pretrained model outperforms them by a large margin. This is not surprising as our pretraining data contain better aligned retrieval supervision signals: reference documents often have supporting evidence for the question while their retrieval training signals are relatively indirect.

In the \textbf{domain adaptation} and \textbf{supervised} settings, our pretrained dense retriever provides a better finetuning initialization and leads to improvement compared with randomly initialized DPR models. Another surprising result is that our  pretrained dense retriever even outperforms some DPR domain adaptation models. For example, on the TriviaQA testing set, our pretrained DPR model achieves 72.2\% top-20 and 81.3\% top-100 accuracy while the DPR-NQ model obtains 69.7\% and 79.2\%  respectively. This indicates that our pretrained dense retriever can generalize well even without using any annotated QA instances.

All the results demonstrate the usefulness and generalization of our pretrained dense retriever for open-domain QA tasks.

\subsection{End-to-End QA performance}

We now examine how our pretrained retriever and reader improve the end-to-end QA performance, measured in exact match (EM). The results are shown in Table \ref{tbl:qa_performance}, from which we make the following observations. (i) Surprisingly, our fully-unsupervised system (pretrained retriever + pretrained reader) shows a certain level of open-domain QA ability (see row \#3). For example, on TriviaQA, our fully-unsupervised system can answer around 25\% of questions correctly. 
(ii) Compared to the system with BM25 retriever (row \#4), the one with our pretrained dense retriever (line \#5) retrieves more relevant passages, leading to better QA performance. 
(iii) Initializing either the retriever or the reader from our pretrained checkpoint can lead to further improvement (rows \#6-\#8). For example, on the TriviaQA and WebQuestion datasets, our entire pipeline pretrain leads to about 4\% absolute gain in terms of EM. Note that on the WebQuestion dataset, all the DPR models perform worse than REALM, this is because of the limited training data of WebQuestion. The issue can be easily solved by adding \textit{Multi} datasets for finetuning according to \cite{karpukhin2020dense}.

\subsection{Computational Resource Comparison}
In addition to the performance gain, another benefit of \model is its training scalability. We compare the \model pretraining with ORQA and REALM in terms of computational resources they use in Table \ref{tbl:computation}. As can be seen, \model only requires reasonable GPU computational resources, which could be normally conducted on an academic-level computational platform. On the contrary, due to the lack of direct retrieval supervision, ORQA and REALM often needs more computational resources and requires more training steps to converge.

\begin{table}[t]
\resizebox{\linewidth}{!}{%
\begin{tabular}{l|ccc|ccc}
\hline
 & \multicolumn{3}{c|}{GPU} & \multicolumn{3}{c}{TPU} \\ \hline
 & \#cards & \begin{tabular}[c]{@{}c@{}}batch \\ size\end{tabular} & \begin{tabular}[c]{@{}c@{}}Train \\ steps\end{tabular} & \#cards & \begin{tabular}[c]{@{}c@{}}batch \\ size\end{tabular} & \begin{tabular}[c]{@{}c@{}}Train \\ steps\end{tabular} \\ \hline
ORQA & 128 & 4096 & 100K & - & 4096 & 100K \\ \hline
REALM & 240 & - & 100K & 64 & 512 & 200K \\ \hline
\model & 8 & 56 & 20K & - & - & - \\ \hline
\end{tabular}
}
\caption{Computational resource comparison between different retriever pretraining  methods. Our \model provides more direct retrieval pretraining signals, thus leading to fast converge. ORQA and REALM GPU setups are from \cite{SachanPSKPHC20} and TPU setups are from their original papers. }
\label{tbl:computation}
\end{table}

\section{Conclusion}
This paper proposes an effective approach for pretraining open-domain QA systems. Specifically, we automatically construct three million pseudo question-answer-context triplets from Wikipedia that align well with open-domain QA tasks. Extensive experiments show that
pretraining a widely-used open-domain QA model (DPR) on our constructed data achieves promising performance gain in both retrieval and QA accuracies. Future work includes exploring the effectiveness of the constructed data on more open-domain QA models (e.g., REALM) and training strategies (e.g., joint optimizing the retriever and reader).

\section*{Acknowledgements}
The authors would thank the anonymous reviewers for their insightful comments and suggestions. The authors would also thank the colleagues in Tencent AI Lab for their internal discussions and feedback. 


\appendix

\section{Appendix}
\label{sec:appendix}
\begin{table}[h]
\centering
\resizebox{\linewidth}{!}{%
\begin{tabular}{l|l}
\hline
\textbf{NER Type} & \textbf{Candidate Question Phrases} \\ \hline
CARDINAL & "what", \\ \hline
DATE & \begin{tabular}[c]{@{}l@{}}"when","what time", \\ "what date",\end{tabular} \\ \hline
EVENT & \begin{tabular}[c]{@{}l@{}}"what event","what",\\ "which event",\end{tabular} \\ \hline
FAC & "where","what buildings", \\ \hline
GPE & "where", "what country", \\ \hline
LANGUAGE & "what language","which language", \\ \hline
LAW & "which law","what law", \\ \hline
LOC & \begin{tabular}[c]{@{}l@{}}"where", "what location", \\ "which place", "what place",\end{tabular} \\ \hline
MONEY & "how much money","how much", \\ \hline
NORP & "what", "what groups", "where", \\ \hline
ORDINAL & "what rank","what", \\ \hline
ORG & \begin{tabular}[c]{@{}l@{}}"which organization", \\ "what organization", "what",\end{tabular} \\ \hline
PERCENT & "what percent", "what percentage", \\ \hline
PERSON & "who", "which person", \\ \hline
PRODUCT & "what", "what product", \\ \hline
QUANTITY & "how many", "how much", \\ \hline
TIME & "when", "what time", \\ \hline
WORK\_OF\_ART & "what", "what title" \\ \hline
\end{tabular}
}
\caption{Question phrase replacement rules for different types of entities.}
\label{tbl:question_word}
\end{table}



\end{document}